\documentclass{article} 
\usepackage{iclr2026_preprint,times}

\usepackage{amsmath,amsfonts,bm}

\def\eqref#1{equation~\ref{#1}}

\def\1{\bm{1}}

\DeclareMathAlphabet{\mathsfit}{\encodingdefault}{\sfdefault}{m}{sl}
\SetMathAlphabet{\mathsfit}{bold}{\encodingdefault}{\sfdefault}{bx}{n}

\PassOptionsToPackage{numbers,compress}{natbib}
\usepackage{graphicx}
\usepackage{amsthm}
\theoremstyle{plain}

\theoremstyle{definition}

\usepackage{wrapfig}
\usepackage{algorithm}
\usepackage{algorithmic}
\usepackage{tabularray}
\usepackage{amssymb}
\usepackage{subcaption}
\usepackage{multirow}
\usepackage{arydshln}
\usepackage{tabularx}
\usepackage[utf8]{inputenc} 
\usepackage[T1]{fontenc}    
\usepackage{hyperref}       
\usepackage{url}            
\usepackage{booktabs}       
\usepackage{amsfonts}       
\usepackage{nicefrac}       
\usepackage{microtype}      
\usepackage{xcolor}         
\usepackage{arydshln}
\usepackage{tikz}
\usepackage{pgfplots}
\usepackage{todonotes} 

\title{CHASE: Channel-Aligned Structure Exploitation for Geometry-Aware Model Engineering}

\author{
Wei Wang\textsuperscript{*} \\
Futurewei Technologies\\
San Jose, CA 95131, USA \\
\texttt{rickweiwang@futurewei.com} \\
\And 
Wei Jiang\textsuperscript{*} \\
Futurewei Technologies\\
San Jose, CA 95131, USA \\
\texttt{wjiang@futurewei.com} \\
\And
Ziran Liu \\
Shanghai Institute for Mathematics and Interdisciplinary Sciences (SIMIS)\\
Shanghai, 200433, China\\
\texttt{zliu@simis.cn}
}

\iclrfinalcopy 
\begin{document}

\maketitle

\begingroup
\renewcommand{\thefootnote}{*}
\footnotetext{Equal contribution.}
\endgroup

\begin{abstract}
Geometric and Spectral Alignment (GSA) characterizes trained networks through spectral concentration, physical-channel alignment, support structure, and changes in singular bases. In this paper, we propose CHASE (\emph{Channel-Aligned Structure Exploitation}) to use these structures in practical model design. CHASE covers six applications across model modification, reconfiguration, and compression. CORA, COEC, and CORAM apply GSA to parameter-efficient finetuning, structured-pruning compensation, and model merging. We further develop three new methods. CAGA  uses GSA to identify multi-head attention heads that can share a KV representation and constructs the shared key and value heads through geometric alignment and low-rank subspace extraction. SAKV uses GSA to determine which adjacent layers can share a low-rank KV-cache representation and the retained rank for each layer group. CAPS uses GSA spectral structure to group output neurons and selects retained input channels separately for each group. Results from CORA, COEC, and CORAM establish the effectiveness of GSA for adaptation, pruning compensation, and model merging. Experiments on CAGA show that geometric shared-head construction substantially improves MHA-to-GQA conversion, and SAKV and CAPS improve over representative baselines for KV-cache compression and structured pruning. These results show that the structures identified by GSA can be used directly to design methods for a range of model operations.
\end{abstract}

\section{Introduction}
\label{sec:intro}

Deep neural networks often develop structural regularities that are not directly specified by their architectures or training objectives. A small number of channels can dominate activation magnitude and quantization error \cite{llmint8,smoothquant,awq,massiveact}. Factual knowledge and even broad behaviors can be localized to a small number of parameters or directions \cite{rome,memit,refusal}. Large portions of a trained model, including channels, heads, layers, and experts, can often be removed or shared with very small loss \cite{shortgpt,expertdrop}. These observations have motivated separate lines of work in quantization, model editing, pruning, efficient adaptation, and sparse architectures. These observations indicate that trained networks share a common underlying organization, with useful computation concentrated in a relatively small set of directions and interactions despite the much larger parameter space.

Geometric and Spectral Alignment (GSA) \cite{gsa1,gsa2} provides a way to characterize this organization from two perspectives. The first is spectral. If most of a layer transformation is concentrated in a small number of singular directions, this gives a natural explanation for why many parameters or channels can be removed with limited loss and why low-rank adaptation can be effective. The second is relational. If the dominant directions of adjacent layers repeatedly align through the same physical channels, then some channels become strongly associated with particular parts of the computation while others are reused across several parts. This provides a possible structural basis for localized knowledge, shared computation, hub-like channels, and redundancy across model components. GSA further studies how this organization changes when a model is modified, which is directly related to why some low-dimensional updates preserve a pretrained model better than unconstrained changes.

In particular, GSA-I \cite{gsa1} characterizes how the energy of a layer transformation is distributed across its singular directions and introduces an effective-rank window that measures how many directions are needed to capture a given fraction of the spectral energy. A small effective-rank window means that most of the transformation is concentrated in a relatively low-dimensional subspace. This provides a direct structural signal for low-rank adaptation, model compression, and layerwise rank allocation.

GSA-II \cite{gsa2} then studies how the low-dimensional subspaces identified by the effective-rank windows align across adjacent layers. For two consecutive layers, it compares the retained output subspace of the first layer with the retained input subspace of the next and expresses this relationship in the original physical channel coordinates through a Physical Alignment Matrix. The rows of this matrix are grouped according to the retained spectral components. For each row group, GSA-II identifies the physical channels that carry most of the corresponding alignment energy. These channels form the active support of that group.

The resulting support pattern shows how the retained inter-layer structure is distributed across physical channels. Some channels are mainly associated with one group, while others are shared by several groups. GSA-II separates this structure into core, overlap, and noise components. The core contains strong connections associated with individual groups. The overlap captures strong connections carried by channels that are shared across groups. The noise contains the remaining weakly structured entries. Channels that appear in the active supports of multiple groups are identified as hub channels. GSA-II also defines margins that measure how clearly the active supports and their overlaps are separated and when this channel structure remains stable under perturbation.

The structure identified by GSA may explain several of the observations described above. Hub channels, which participate in the active supports of multiple groups, may correspond to the small set of high-impact channels observed in activation and quantization studies. Channels associated mainly with individual groups may help explain why certain knowledge or behaviors can be localized to restricted parts of a network. Components with weak or redundant support are natural candidates for removal or sharing. The same structural information can also guide model design. Spectral concentration indicates how much of a layer transformation needs to be retained, while channel support identifies which components are local, shared, or weakly represented. The coherent singular-frame result further provides a structured way to modify a pretrained weight without strongly disturbing its original geometry. These properties are directly relevant to parameter-efficient finetuning (PEFT), structured pruning and pruning compensation, model merging, attention reconfiguration, and KV-cache compression. 

Several applications from our prior work have already demonstrated practical uses of GSA. CORA \cite{cora} uses coherent changes in the left and right singular bases for PEFT. COEC \cite{coec} compensates for changes in the input basis, output basis, and singular values introduced by structured pruning. CORAM \cite{coram} represents finetuned models through rotational, spectral, and right-subspace factors and merges these factors separately.

In this paper, we propose \textbf{CHASE} (\emph{Channel-Aligned Structure Exploitation}) as an application framework for using GSA measurements in model design. We include CORA, COEC, and CORAM as established applications and develop three new methods. CAGA (\emph{Coherent Alignment for Grouped Attention}) uses GSA to identify multi-head attention (MHA) heads that can share a KV representation and constructs the shared key and value heads through geometric alignment and low-rank subspace extraction. SAKV (\emph{Spectrally Aligned KV-Cache Compression}) uses GSA to determine which adjacent layers can share a low-rank KV-cache representation and the retained rank for each layer group. CAPS (\emph{Channel-Aligned Pruning Selection}) uses GSA spectral structure to group output neurons and selects retained input channels separately for each group.


Different applications use different GSA measurements. CORA, COEC, and CORAM use singular-basis and spectral changes when pretrained weights are modified, compensated, or merged. CAGA uses physical-channel support overlap together with retained spectral and right-subspace similarity to group attention heads. SAKV extends the effective-rank criterion to shared KV representations, while CAPS uses the effective-rank spectral directions to organize output neurons. CHASE links each model operation to the GSA quantities used by that operation.

Our main contributions are as follows.
\begin{itemize}

\item We propose CHASE, an application framework that uses the spectral, physical-channel, and singular-basis measurements of GSA to guide model modification, reconfiguration, and compression.

\item We present six CHASE applications. CORA, COEC, and CORAM cover PEFT, structured-pruning compensation, and model merging. We further develop CAGA for multi-head attention grouping, SAKV for KV-cache compression, and CAPS for structured-pruning  selection, 
and evaluate the three new methods against representative baselines for their respective applications.
\end{itemize}

\section{Geometric and Spectral Structures Used in CHASE}
\label{sec:gsa-signals}

CHASE builds on several structures introduced in GSA-I and GSA-II. Three types of GSA information are used directly in the applications considered here. Spectral concentration and effective rank are used in SAKV and CAPS. Physical-channel support, together with retained spectral and subspace similarity, is used in CAGA. Coherent singular-frame motion is used in CORA, COEC, and CORAM. We also summarize the support-stability quantities from GSA-II, which characterize the reliability of the extracted support structure. The complete theoretical development, including the geometric formulation, assumptions, stability conditions, and proofs, is given in GSA-I and GSA-II~\cite{gsa1,gsa2}.

\subsection{Spectral concentration and effective rank}

For a matrix $W\in\mathbb{R}^{m\times n}$ with singular values
$\sigma_1\geq\cdots\geq\sigma_d$, where $d=\min(m,n)$, define the fraction of spectral energy captured by the first $R$ singular directions as
\begin{equation}
\mathcal E_R(W)
=
\frac{\sum_{i=1}^{R}\sigma_i^2}
     {\sum_{i=1}^{d}\sigma_i^2}.
\end{equation}
GSA-I \cite{gsa1} gives the corresponding effective-rank window as
\begin{equation}
R_{\varepsilon}(W)
=
\min
\left\{
R:
\mathcal E_R(W)\geq 1-\varepsilon
\right\}.
\label{eq:effective-rank}
\end{equation}
The quantity $R_{\varepsilon}(W)$ measures how many singular directions are needed to retain a given fraction of the spectral energy. A small ratio $R_{\varepsilon}/d$ means that most of the transformation is concentrated in a low-dimensional subspace, while a larger ratio indicates that the spectral energy is spread across more directions. The geometric formulation and theoretical analysis underlying this quantity are described in detail in GSA-I.

CHASE uses the effective-rank criterion in SAKV and CAPS. SAKV extends the retained-spectral-energy criterion to grouped KV representations and uses it to determine the rank required by each layer group. CAPS uses \(R_\varepsilon(W)\) to determine the singular directions used to organize output neurons into groups. Spectral energy does not directly measure task importance, so calibration information is added when an application requires task-dependent selection.

\subsection{Inter-layer alignment in physical channels}

The effective-rank window identifies a retained low-dimensional subspace for each layer. GSA-II studies how these subspaces align across consecutive layers \cite{gsa2}. For two adjacent layer operators, it compares the retained output subspace of the first layer with the retained input subspace of the next layer. The resulting relationship is then expressed in the original physical channel coordinates of the model through the Physical Alignment Matrix, which we denote by $M_{\mathrm{phy}}$. 

The rows of $M_{\mathrm{phy}}$ are grouped according to the retained spectral components. For each row group $i$, GSA-II identifies a set of physical channels $\mathcal C_i$ that carries most of the corresponding alignment energy. The set $\mathcal C_i$ is the active support of that group. These active supports reveal how the retained inter-layer structure is distributed across physical channels. Some channels are mainly associated with one group, while others appear in the supports of several groups. Based on this support pattern, GSA-II decomposes the Physical Alignment Matrix as
\begin{equation}
 M_{\mathrm{phy}}
=
M_{\mathrm{core}}
+
M_{\mathrm{overlap}}
+
M_{\mathrm{noise}}.
\label{eq:pam-decomp}
\end{equation}
The core contains strong connections associated with individual groups. The overlap contains strong connections carried by channels shared across groups. The noise contains the remaining weakly structured entries. The number of active supports containing a physical channel $c$ is
\begin{equation}
d(c)
=
\left|
\left\{
i:
c\in\mathcal C_i
\right\}
\right|.
\label{eq:hub-degree}
\end{equation}
Channels with $d(c)\geq 2$ are shared-support or hub channels.

The active supports are used directly in CAGA. For candidate attention-head pairs, CAGA combines physical-channel support overlap with retained singular-value similarity and right-subspace similarity to determine which heads are suitable for sharing.

\subsection{Stability of the channel structure}

The active supports in the Physical Alignment Matrix are determined from the energy carried by each physical channel. For a row group $\mathcal G_i$, GSA-II assigns every channel $c$ the column-energy score
\begin{equation}
q_i(c)
=
\left\|
M_{\mathrm{phy}}[\mathcal G_i,\{c\}]
\right\|_2^2 .\label{eq:channel-energy}
\end{equation}
The channels are ranked according to $q_i(c)$. The active support $\mathcal C_i$ is then formed either by retaining the smallest set of highest-energy channels whose summed energy reaches a given fraction of the total energy of the group, or by retaining a fixed number of the highest-energy channels \cite{gsa2}.

GSA-II measures how stable this support is through the active-column gap. This gap compares the weakest channel inside $\mathcal C_i$ with the strongest channel outside it. A large gap means that the active support is clearly separated from the remaining channels. A small gap means that channels near the support boundary can change their ordering under a small perturbation.

The theory also considers the overlap between the active supports of different row groups. For two groups, some channels may be shared by both supports, while others belong only to one of them. GSA-II defines pairwise overlap quantities and corresponding margins that measure how clearly the shared channels are separated from the group-specific channels. Together with the residual noise level, these quantities characterize how well the core, overlap, and noise structure is separated.

GSA-II combines the active-column gaps, pairwise overlap margins, and residual bounds into a static certificate radius $r_{\mathrm{cert}}$. For a fixed row grouping and active-support rule, the theory shows that when the perturbation of the Physical Alignment Matrix satisfies the required bounds, the active supports remain unchanged. The pairwise support relations, hub channels, and the core--overlap--noise masks remain unchanged as well. The complete definition of $r_{\mathrm{cert}}$, the perturbation conditions, and the proof are given in GSA-II \cite{gsa2}.

The active-column gaps, pairwise overlap margins, and certificate characterize how clearly the extracted support structure is separated and how stable it remains under perturbation. These quantities provide the theoretical basis for interpreting the reliability of the channel structure identified by GSA-II. The certificate is derived for fixed-dimensional perturbations, while CHASE
uses the corresponding measured structure to guide dimension-changing model operations.

\subsection{Coherent singular-frame motion}

GSA-II also studies how a pretrained weight can change while remaining close to its original geometry. Given a pretrained weight 
\begin{equation}
W_0
=
U_0\Sigma_0V_0^\top, \nonumber
\end{equation}
a modified weight can be represented in the singular coordinates of $W_0$ as
\begin{equation}
W
=
U_0Q_U\Sigma Q_V^\top V_0^\top,
\label{eq:base-frame}
\end{equation}
where $Q_U$ and $Q_V$ are changes to the left and right singular bases, and $\Sigma$ is the spectral change.

GSA-II analyzes the perturbation of $W_0$ in the singular-vector coordinates of the pretrained weight. Under the low-disruption conditions developed in the theory, small model changes constrain the relative rotation between the left and right singular frames. In particular, low-cost changes favor coherent motion in which $Q_U$ and $Q_V$ remain close rather than rotating independently. The full perturbation bounds and their assumptions are given in GSA-II \cite{gsa2}.

The analysis of coherent changes in the left and right singular bases provides a useful way to parameterize modifications of a pretrained weight. Rather than representing the update as an arbitrary matrix, the change can be separated into rotations of the pretrained singular bases and changes to the singular values. This separation is useful in applications that need to modify, compensate, or combine pretrained weights while preserving their existing geometric structure.

Table~\ref{tab:signal-map} summarizes the GSA quantities discussed above and their use in CHASE. Effective rank provides the spectral dimensionality used by SAKV and CAPS. CAGA uses physical-channel support overlap together with spectral and right-subspace similarity. CORA, COEC, and CORAM use changes in singular bases and singular values when modifying, compensating, or combining pretrained weights.

\begin{table}[t]
\centering
\caption{GSA quantities and their use in CHASE.}
\label{tab:signal-map}
\small
\setlength{\tabcolsep}{4pt}
\begin{tabular}{p{0.24\linewidth} p{0.4\linewidth} p{0.29\linewidth}}
\toprule
\textbf{GSA quantity} &
\textbf{What it measures} &
\textbf{Use in CHASE} \\
\midrule

Effective-rank window
$R_{\varepsilon}$ &
Number of leading singular directions needed to retain most of the spectral
energy &
SAKV rank selection, CAPS output-neuron grouping \\
\addlinespace
Active supports
$\{\mathcal{C}_i\}$ &
Which physical channels carry the retained inter-layer structure for each
spectral group &
CAGA head grouping through support overlap \\
\addlinespace
Retained spectral and
right-subspace similarity &
Similarity of retained singular values and input subspaces between model
components &
CAGA head grouping \\
\addlinespace
Basis rotations
$Q_U$, $Q_V$ and spectral change &
Changes in left and right singular bases and singular values relative to
pretrained weight &
CORA adaptation, COEC compensation, CORAM merging \\
\addlinespace
Active-column gap, pairwise
overlap margin, $r_{\mathrm{cert}}$ &
Separation and stability of the extracted physical-channel supports &
GSA support stability analysis\\

\bottomrule
\end{tabular}
\end{table}

\section{CHANNEL-ALIGNED STRUCTURE EXPLOITATION}
\label{sec:model-modification}

This section presents six CHASE applications spanning model modification, reconfiguration, and compression. CORA (\emph{Coherent Orthogonal Rotation Adaptation}) uses coupled rotations of the left and right singular bases for parameter-efficient finetuning. COEC (\emph{Calibrated Orthogonal-Equivalence Compensation}) corrects the changes to the singular bases and singular values introduced by structured pruning. CORAM (\emph{Coherent Rotation-Aware Model Merging}) separates rotational, spectral, and subspace changes when combining finetuned models. CAGA (\emph{Coherent Alignment for Grouped Attention}) uses GSA to identify MHA heads that can share a KV representation and constructs the shared key and value heads through geometric alignment and low-rank subspace extraction. 
SAKV (\emph{Spectrally Aligned KV-Cache Compression}) uses GSA to determine which adjacent layers can share a low-rank KV-cache representation and the retained rank for each layer group. CAPS (\emph{Channel-Aligned Pruning Selection}) uses GSA spectral structure to group output neurons and selects retained input channels separately for each group. The six applications cover parameter adaptation, pruning compensation, model merging, attention reconfiguration, KV-cache compression, and pruning-mask selection. In each case, the model operation is determined from the spectral, physical-channel, or singular-basis structure described in Section~\ref{sec:gsa-signals}.

\subsection{CORA: Coherent Singular-Basis Rotation for PEFT}
\label{sec:cora}

LoRA-style PEFT methods \cite{lora} represent the change to a pretrained weight as an additive low-rank matrix,
\begin{equation}
\widetilde W = W_0 + BA.
\nonumber
\end{equation}
Approaches like DoRA \cite{dora} and PiSSA \cite{pissa} refine the low-rank parameterization or its initialization, but keep the additive modification of the pretrained weight. This representation reduces the number of trainable parameters, but it does not explicitly separate changes to the left singular basis, the right singular basis, and the singular values of $W_0$.

Orthogonal finetuning methods like OFT \cite{oft}  parameterize adaptation through rotations, which makes changes to the singular bases explicit. However, a rotation-only update cannot independently adjust the singular values. CORA \cite{cora} addresses this issue by combining coherent rotations of the two singular bases with a separate spectral change.

\paragraph{Method.}
As described in Section~\ref{sec:gsa-signals}, GSA-II represents a modified pretrained weight $W_0$ in its singular-vector coordinates as Eq.~\ref{eq:base-frame}. Under its low-disruption conditions, the relative rotation between the left and right singular bases is constrained to be small. CORA enforces the strongest form of this coherence by setting $Q_U=Q_V=Q$, while allowing the singular values to change separately.

To keep the rotation compact, CORA applies coherent rotation to low-rank row slices of the pretrained weight. For a slice
\begin{equation}
W_i
=
U_i\Sigma_iV_i^\top,\nonumber
\end{equation}
the adapted slice is parameterized as
\begin{equation}
\widetilde W_i = U_iQ_i
(\Sigma_i+\Delta\Sigma_i)
Q_i^\top V_i^\top.
\label{eq:cora}
\end{equation}
The same orthogonal matrix $Q_i$ rotates the left and right singular bases, while $\Delta\Sigma_i$ controls the change in singular values. Each $Q_i$ is parameterized by the Cayley transform of a skew-symmetric matrix, which keeps the rotation orthogonal throughout finetuning.

CORA applies this parameterization to a rank-$r$ reconstruction of the pretrained weight and keeps the remaining residual fixed. A shared layerwise spectral parameter provides an additional low-cost adjustment to the singular values. For an $m\times k$ layer with full $r\times r$ Cayley blocks, the rotations require approximately $\frac{1}{2}m(r-1)$ trainable parameters. For square projections at the same rank, this is approximately one quarter of the parameter count required by LoRA.

\paragraph{Results.}
On LLaMA-2-7B commonsense reasoning, CORA with rank $16$ uses 6.6M trainable parameters and achieves an average accuracy of 82.16\%. On LLaMA-3-8B, the rank-$16$ configuration uses 10.3M trainable parameters and achieves 86.65\%. On HumanEval, CORA with 10.3M trainable parameters reaches 48.2 Pass@1 on LLaMA-3-8B and 40.2 on Mistral-7B. CORA matches or exceeds the evaluated LoRA and SVD-based PEFT baselines while using much fewer trainable parameters.

The mathematical-reasoning experiments also show that restricting adaptation to the leading singular subspace is not equally effective for all tasks. Using the full pretrained weight recovers part of the performance loss observed with the rank-restricted source. The effective-rank window measures spectral concentration, while task-relevant information can also remain in lower-energy directions.

\subsection{COEC: Two-Sided Geometric Compensation for Structured Pruning}
\label{sec:coec}

Training-free structured pruning usually contains two parts. A selection rule chooses which columns to retain, and a compensation step adjusts the retained weight after column removal. Representative methods include Wanda-sp \cite{wanda}, FLAP \cite{flap}, and RCPU \cite{rcpu}. Existing compensation either leaves the retained weight unchanged, adds a bias, or applies a one-sided rotation. These corrections cannot recover changes in both the input and output singular bases caused by column removal, which limits how closely the pruned layer can reproduce the original transformation.

\paragraph{Method.}
The GSA representation in Section~\ref{sec:gsa-signals} separates a weight transformation into its input singular basis, output singular basis, and singular values. After columns are removed, the retained matrix generally differs from the original weight in all three components. COEC~\cite{coec} corrects both the left and the right singular bases through left and right rotations and adjusts the singular values separately.

Given the retained weight $W_K$, COEC constructs
\begin{equation}
\widetilde W
=
Q_{\ell} W_K^{(S)} Q_r,
\label{eq:coec}
\end{equation}
where $Q_{\ell}$ rotates the output basis, $Q_r$ rotates the retained input basis, and $W_K^{(S)}$ adjusts the singular values of $W_K$ individually.

The left rotation is obtained by an orthogonal Procrustes solution that aligns the output of the retained weight with the original layer output $Y=W_0X$, where $X$ contains input activations collected from a small calibration set. The right rotation corrects the input basis changed by column removal and is optimized under an orthogonality constraint on a reduced Stiefel manifold.

For the spectral correction, the singular values are adjusted to reduce the reconstruction error between the compensated pruned layer and the original outputs $Y$. The regularization strength is selected separately for each layer by generalized cross-validation. COEC alternates the left rotation, right rotation, and singular-value rescaling until convergence. All updates use the collected calibration statistics and require no backpropagation through or retraining of the language model.

\paragraph{Results.}
At 30\% structured sparsity, COEC reduces WikiText-2 perplexity for every evaluated model under Wanda-sp, FLAP, and RCPU selections and improves zero-shot accuracy in most settings. Under Wanda-sp selection on Qwen2.5-32B, perplexity decreases from 13.90 to 9.64. The analysis in COEC also shows that column pruning changes the measured input directions by about $22^\circ$ on average in the evaluated setting. The alignment component reduces the remaining mismatch after compensation.

COEC modifies only the retained weights and does not introduce additional parameters or computation at inference time. The parameter, memory, and FLOP reductions obtained from structured pruning are  preserved after compensation.

\subsection{CORAM: Geometry-Compatible Aggregation for Model Merging}
\label{sec:coram}

Model merging combines independently finetuned models without joint training or access to their training data. Methods such as Task Arithmetic \cite{taskarithmetic}, TIES \cite{ties}, and DARE \cite{dare} operate directly on weight differences from a common base model. This treats the update as a vector in Euclidean weight space and does not separate changes to the singular bases from changes to the singular values. OrthoMerge \cite{orthomerge} introduces geometric merging through orthogonal transformations, but a single orthogonal transformation for an entire weight matrix cannot represent spectral changes.

\paragraph{Method.}
The GSA representation in Section~\ref{sec:gsa-signals} separates changes to the singular bases from changes to the singular values of a pretrained weight. When several models are finetuned from the same pretrained model, their changes can be expressed relative to the same pretrained weight. CORAM~\cite{coram} uses this representation to merge the changes in the singular bases and changes in the singular values separately.

For a pretrained weight matrix $W_0$, CORAM partitions its rows into slices indexed by $g$. The SVD of the $g$-th slice is
\begin{equation}
W_0^{(g)}
=
U_0^{(g)} S_0^{(g)} {V_0^{(g)}}^\top .\nonumber
\end{equation}
For each finetuned model $i$, the corresponding slice is
\begin{equation}
W_i^{(g)}
=
U_i^{(g)} S_i^{(g)} {V_i^{(g)}}^\top ,\nonumber
\end{equation}
where $i=1,\ldots,N$ indexes the models being merged.

CORAM describes how $W_i^{(g)}$ differs from $W_0^{(g)}$ using three quantities. The rotation
\begin{equation}
Q_i^{(g)}
=
\operatorname{polar}
\left(
{U_0^{(g)}}^\top U_i^{(g)}
\right)\nonumber
\end{equation}
captures the change in the left singular basis, where
$\operatorname{polar}(A)=UV^\top$ for the SVD
$A=U\Sigma V^\top$. The relative spectral shift
\begin{equation}
dS_i^{(g)}
=
S_i^{(g)}
{S_0^{(g)}}^{-1}
-I\nonumber
\end{equation}
captures the change in singular values. The right singular basis $V_i^{(g)}$ gives the corresponding input-side subspace of the finetuned slice.

These three quantities have different geometric constraints and are merged separately. The rotations are averaged on the special orthogonal group,
\begin{equation}
\bar Q^{(g)}
=
\exp\left(
\frac{1}{N}
\sum_{i=1}^{N}
\log Q_i^{(g)}
\right).\nonumber
\end{equation}
The relative spectral shifts are averaged directly,
\begin{equation}
\overline{dS}^{(g)}
=
\frac{1}{N}
\sum_{i=1}^{N}
dS_i^{(g)},\nonumber
\end{equation}
while the right singular bases are averaged and projected back to the Stiefel manifold,
\begin{equation}
\bar V^{(g)}
=
\operatorname{polar}
\left(
\frac{1}{N}
\sum_{i=1}^{N}
V_i^{(g)}
\right).\nonumber
\end{equation}
The merged slice is reconstructed using the pretrained left basis and singular values,
\begin{equation}
W_{\mathrm{merge}}^{(g)}
=
\left(
U_0^{(g)}\bar Q^{(g)}
\right)
S_0^{(g)}
\left(
I+\overline{dS}^{(g)}
\right)
(\bar V^{(g)})^{\top}.
\label{eq:coram}
\end{equation}
The reconstructed slices are stacked to obtain the merged weight matrix \(W_{\mathrm{merge}}\).

Averaging the factors from several finetuned models can partially cancel their individual changes, making the merged update
$\|W_{\mathrm{merge}}-W_0\|$
smaller than a typical finetuned-model update
$\|W_i-W_0\|$.
CORAM compensates for this reduction by scaling the merged update,
\begin{equation}
W^{*}
=
W_0
+
\lambda
\left(
W_{\mathrm{merge}}-W_0
\right),
\qquad
\lambda=\kappa\widehat c .
\nonumber
\end{equation}
The scale $\widehat c$ is computed by comparing the typical magnitude of the individual finetuning updates with the magnitude of the merged update,
\begin{equation}
\widehat c
=
\frac{
\sqrt{\frac{1}{N}\sum_{i=1}^{N}\|W_i-W_0\|^2}
}{
\|W_{\mathrm{merge}}-W_0\|
}.
\nonumber
\end{equation}
The factor $\kappa$ controls how much of this lost update magnitude is restored. CORAM compares the largest and smallest relative update magnitudes among the finetuned models,
\begin{equation}
D
=
\frac{\max_i r_i}{\min_i r_i},
\qquad
r_i
=
\frac{\|W_i-W_0\|}{\|W_0\|}.
\nonumber
\end{equation}
When the finetuned models change the pretrained model by similar amounts, a larger $\kappa$ is used. When their update magnitudes differ greatly, a smaller $\kappa$ is used. This determines $\lambda$ directly from the model weights without testing multiple values of $\lambda$ on downstream tasks.

\paragraph{Results.}
CORAM is evaluated on Llama-3.1-8B, Llama-3.2-3B, Qwen2.5-VL-7B, and Gemma-2-9B, covering both language and vision-language model merging. It improves over OrthoMerge \cite{orthomerge} by 0.25--1.35 points and performs as well as or better than Task Arithmetic \cite{taskarithmetic}, TIES \cite{ties}, and DARE \cite{dare} in the reported experiments \cite{coram}. The value of $\lambda$ chosen from the model updates is within 0.72 points of the best value found by evaluating multiple choices of $\lambda$.



\subsection{CAGA: Coherent Alignment for Grouped Attention}
\label{sec:caga}

Multi-head attention (MHA) maintains separate key and value projections for each query head, while grouped-query attention (GQA) reduces KV-cache memory by allowing several query heads to share one key/value head \cite{gqa}. Converting a pretrained MHA model to GQA requires two decisions. The first is which query heads should share a key/value head. The second is how the original key and value projections should be replaced by a single shared pair. Existing conversion methods use neighboring heads, activation similarity, or head alignment to determine the groups ~\cite{gqa,asymgqa,mha2gqa_align}, and the shared projections are commonly obtained by averaging or aligned averaging.

Directly averaging the key and value projections of several heads can lose information because different heads may represent related features in different coordinate systems. CAGA uses GSA to identify heads with compatible spectral and subspace structure and then constructs their shared key and value projections accordingly. For the keys, we align each head with a shared key projection using RoPE-compatible transformations and absorb the corresponding correction into the query projection. For the values, we extract a shared low-rank subspace from the grouped value projections and absorb the head-specific factors into the output projection. The converted layer retains the standard GQA structure and requires no additional operations during inference.

\paragraph{Method.}
Let
\(
W_{Q,h},W_{K,h},W_{V,h}\in\mathbb{R}^{d_h\times d}
\)
denote the query, key, and value projections of head $h$, and let
\(W_{O,h}\in\mathbb{R}^{d\times d_h}
\)
denote the corresponding block of the output projection. Let $H$ denote the number of query heads in the original MHA layer and $G$ the number of key/value heads after GQA conversion. Each GQA group  contains $m=H/G$ query heads that share one key projection $\overline W_K$ and one value projection $\overline W_V$.

We first characterize candidate head groups using the GSA measurements in
Sec.~\ref{sec:gsa-signals}. For a pair of heads $a$ and $b$, we compute their
physical-channel support overlap, retained singular-value similarity, and
retained right-subspace similarity. These quantities have different numerical
ranges, so each pairwise quantity is converted to its percentile rank among
all head pairs in the same layer. Let
$r_{\mathrm{sup}}(a,b)$,
$r_{\mathrm{spec}}(a,b)$, and
$r_{\mathrm{sub}}(a,b)$ denote the resulting normalized distances. CAGA uses
\begin{equation}
D_{ab}^{\mathrm{CAGA}}
=
\eta_1 r_{\mathrm{sup}}(a,b)
+
\eta_2 r_{\mathrm{spec}}(a,b)
+
\eta_3 r_{\mathrm{sub}}(a,b),\nonumber
\end{equation}
with $\eta_1=\eta_2=\eta_3=1$ by default. The heads are grouped to equal-size groups by minimizing the total distance among heads within groups. When $m=2$, we solve this as a minimum-cost pairing problem. For larger groups, we use balanced clustering. The grouping determines which heads are considered jointly. CAGA then uses the geometry of the grouped key and value projections to construct their shared representation.

The key projections cannot be aligned with an arbitrary transformation because rotary positional encoding (RoPE) is applied before the attention product. CAGA restricts the alignment of each head to a block-diagonal matrix
\begin{equation}
T_h
=
\operatorname{diag}
\left(
T_{h,1},\ldots,T_{h,d_h/2}
\right),
\qquad
T_{h,j}
=
\begin{bmatrix}
a_{h,j} & -b_{h,j}\\
b_{h,j} & a_{h,j}
\end{bmatrix}.\nonumber
\end{equation}
Each $2\times2$ block applies a scaling and rotation to one RoPE coordinate pair.  This transformation is compatible with the RoPE rotation, which can be absorbed into the corresponding query projection.

CAGA runs the pretrained model on a small calibration set and collects the hidden states entering each attention layer. Let $X\in\mathbb{R}^{d\times N}$ contain the hidden states of the $N$ collected tokens as its columns. For a candidate group $\mathcal H$, CAGA finds a shared key projection and per-head transformations by minimizing
\begin{equation}
\min_{\overline W_K,\{T_h\}}
\sum_{h\in\mathcal H}
\left\|
\left(
W_{K,h}-T_h\overline W_K
\right)
\Sigma_X^{1/2}
\right\|_F^2,\quad \Sigma_X=\frac{1}{N}XX^\top.\nonumber
\end{equation}
The transformation $T_h$ does not introduce an additional operation during inference. We incorporate it into the query projection by setting
\begin{equation}
\widetilde W_{Q,h}=T_h^\top W_{Q,h}.
\label{eq:caga-query-absorb}
\end{equation}
The head then uses $\widetilde W_{Q,h}$ together with the shared key projection $\overline W_K$.

For the values, CAGA constructs one shared value projection for all heads in a group. For the group of $m$ heads $\mathcal H=\{h_1,\ldots,h_m\}$,  we stack their value projections as
\begin{equation}
W_{V,\mathcal H}
=
\begin{bmatrix}
W_{V,h_1}\\
W_{V,h_2}\\
\vdots\\
W_{V,h_m}
\end{bmatrix}
\in\mathbb{R}^{md_h\times d}.
\nonumber
\end{equation}
We then compute the rank-$d_h$ truncated singular value decomposition
\(W_{V,\mathcal H}
\approx
U_{d_h}\Sigma_{d_h}V_{d_h}^{\top}.
\)
This is the optimal rank-$d_h$ approximation of the stacked value projections under the Frobenius norm.

The shared value projection is
\begin{equation}
\overline W_V
=
\Sigma_{d_h}V_{d_h}^{\top}. \nonumber
\end{equation}
The rows of $U_{d_h}$ associated with head $h$ form a matrix $B_h\in\mathbb{R}^{d_h\times d_h}$, giving
\begin{equation}
W_{V,h}
\approx
B_h\overline W_V.
\nonumber
\end{equation}
Rather than applying $B_h$ as an additional operation during inference, we incorporate it into the corresponding output projection,
\begin{equation}
\widetilde W_{O,h}
=
W_{O,h}B_h.
\label{eq:caga-output-absorb}
\end{equation}
Therefore, each group uses one shared value projection $\overline W_V$, while the head-specific information is retained through the modified output projections. Eq.~\ref{eq:caga-query-absorb} and Eq.~\ref{eq:caga-output-absorb} produce a standard GQA layer with one shared key/value head for each group and no additional inference-time transformations.

\paragraph{Results.}
CAGA is evaluated on MHA-to-GQA conversion of Llama-2-7B. We find that the choice of head grouping has a relatively small effect once the shared key and value heads are constructed properly. At $G=8$, after 100M tokens of continued training, CAGA, neighboring-head grouping, and the cosine/value-based grouping of~\cite{mha2gqa_align} differ by only 0.23 WikiText-2 perplexity and 1.19 points on the seven-task zero-shot average. In contrast, changing how the shared key and value heads are constructed produces much larger differences.

At $G=16$, which reduces the number of KV heads by $2\times$, directly averaging the key and value projections gives a WikiText-2 perplexity of 926 and a seven-task zero-shot average of 32.6 before any continued training. With the aligned-key and shared-value construction, perplexity drops to 25.1 and the zero-shot average increases to 44.3. The same construction also gives usable models at $G=8$ after continued training, while all tested methods fail at $G=4$.

At $G=8$, after continued training, CAGA, neighboring-head grouping, and the cosine/value-based grouping of~\cite{mha2gqa_align} give very similar results. Across the 100M- and 500M-token settings, the differences among the three grouping methods remain small, confirming that the construction of the shared key and value heads has a much larger effect than the choice of
head grouping. Detailed comparisons and analysis are provided in Appendix~\ref{app:CAGA}.

\subsection{SAKV: Spectrally Aligned KV-Cache Compression}
\label{sec:sakv}

The key-value (KV) cache grows linearly with both context length and the number of Transformer layers, making cache memory a major bottleneck in long-context inference. Existing post-training methods reduce this cost through token removal or low-rank compression. Palu \cite{palu} compresses
the hidden dimension through low-rank projection, while ShadowKV \cite{shadowkv} stores a low-rank key cache and reconstructs selected KV states during decoding. These methods primarily exploit redundancy within individual layers.

Cross-layer methods additionally exploit cross-layer redundancy. Cross-Layer Attention \cite{cla} shares KV states across adjacent layers but changes the attention architecture and requires training. MiniCache \cite{minicache} merges KV states from adjacent layers according to their similarity. More recently, xKV \cite{xkv} shows that the KV representations of several adjacent layers can share a low-rank representation. It groups a fixed number of consecutive layers and uses the same retained key and value ranks for every group. However, fixed grouping and rank allocation ignore variation in spectral structure across depth. Fixed groups may combine poorly aligned layers or separate layers that could share a basis, while fixed ranks may overallocate some groups and underallocate others. Keys and values can also require different ranks.

SAKV uses the spectral quantities introduced in Sec.~\ref{sec:gsa-signals} to make the grouping and rank decisions from the measured KV structure. Eq.~\ref{eq:effective-rank} defines the smallest rank $R_\varepsilon(W)$ whose leading singular directions retain at least $1-\varepsilon$ of the spectral energy of $W$. SAKV extends this criterion from a single matrix to the KV representations of several adjacent layers. It measures whether those layers can share a low-rank spectral space and how
much rank is needed to reconstruct each layer.

\paragraph{Method.}
SAKV uses a small calibration set $\mathcal D_{\mathrm{cal}}$ containing several long input sequences. The calibration set is used only to determine the layer groups and the retained
key and value ranks. No model parameters are trained or modified.

Consider one sequence $x\in\mathcal D_{\mathrm{cal}}$ of length $L$. Let
\begin{equation}
X_{\ell}^{K},X_{\ell}^{V}
\in
\mathbb{R}^{L\times d}
\nonumber
\end{equation}
denote the key and value representations produced by layer $\ell$ for this sequence. Each row corresponds to one token position and each column to one cache dimension. The same statistics are collected over all sequences in $\mathcal D_{\mathrm{cal}}$ when the layer groups and ranks are selected. We use $X_{\ell}^{\tau}$, with $\tau\in\{K,V\}$, when the same operation
applies to keys and values.

The models considered in this work use rotary positional embeddings (RoPE). RoPE applies a position-dependent rotation to the query and key vectors before their inner products are used to compute attention. The value vectors are not rotated. Following xKV \cite{xkv}, SAKV factorizes the key representations before this rotation and factorizes the value representations directly.

Consider a contiguous group of layers
\(
\mathcal L=\{\ell_1,\ldots,\ell_m\}.
\)
Their cache matrices are concatenated,
\begin{equation}
X_{\mathcal L}^{\tau}
=
\left[
X_{\ell_1}^{\tau},
X_{\ell_2}^{\tau},
\ldots,
X_{\ell_m}^{\tau}
\right]
\in
\mathbb{R}^{L\times md}.\nonumber
\end{equation}
The $L$ rows correspond to the same token positions, while the columns
contain the cache dimensions from all layers in the group. We compute the SVD
\begin{equation}
X_{\mathcal L}^{\tau}
=
U_{\mathcal L}^{\tau}
\Sigma_{\mathcal L}^{\tau}
{V_{\mathcal L}^{\tau}}^{\top}.\nonumber
\end{equation}
For a rank-$r$ representation, SAKV keeps the first $r$ singular
directions. Let
$U_{\mathcal L,r}^{\tau}$,
$\Sigma_{\mathcal L,r}^{\tau}$, and
$V_{\mathcal L,r}^{\tau}$ denote the retained components. Define
\begin{equation}
A_{\mathcal L,r}^{\tau}
=
U_{\mathcal L,r}^{\tau}
\Sigma_{\mathcal L,r}^{\tau}
\in
\mathbb{R}^{L\times r}.\nonumber
\end{equation}
The columns of
${V_{\mathcal L,r}^{\tau}}^{\top}$ correspond to the concatenated cache
dimensions in $X_{\mathcal L}^{\tau}$. We divide them into one block for
each layer,
\begin{equation}
{V_{\mathcal L,r}^{\tau}}^{\top}
=
\left[
B_{\ell_1,r}^{\tau},
B_{\ell_2,r}^{\tau},
\ldots,
B_{\ell_m,r}^{\tau}
\right],
\qquad
B_{\ell,r}^{\tau}\in\mathbb{R}^{r\times d}.
\nonumber
\end{equation}
The rank-$r$ reconstruction of layer $\ell$ is
\begin{equation}
\widehat X_{\ell,r}^{\tau}
=
A_{\mathcal L,r}^{\tau}
B_{\ell,r}^{\tau}.
\label{eq:sakv-reconstruction}
\end{equation}
All layers in $\mathcal L$ share
$A_{\mathcal L,r}^{\tau}$, while each layer retains its own
$B_{\ell,r}^{\tau}$. Layers that share a compact spectral space can be
reconstructed accurately with a smaller $r$.

\paragraph{Reconstruction error and retained rank.}
For each candidate group and rank $r$, SAKV evaluates the reconstruction
over the calibration set $\mathcal D_{\rm cal}$. Let
$X_{\ell,x}^{\tau}$ denote the key or value representation of layer $\ell$
for calibration sequence $x$, and let
$\widehat X_{\ell,r,x}^{\tau}$ denote its rank-$r$ reconstruction from
Eq.~\ref{eq:sakv-reconstruction}. For values, the reconstruction error is
\begin{equation}
e_{\ell,r}^{V}
=
\frac{
\displaystyle
\sum\nolimits_{x\in\mathcal D_{\rm cal}}
\left\|
X_{\ell,x}^{V}
-
\widehat X_{\ell,r,x}^{V}
\right\|_F^2
}{
\displaystyle
\sum\nolimits_{x\in\mathcal D_{\rm cal}}
\left\|
X_{\ell,x}^{V}
\right\|_F^2
}.\nonumber
\end{equation}
For keys, raw Frobenius energy can be dominated by a large
token-independent component. We therefore measure the reconstruction error
after RoPE and remove the component shared across token positions. For a
sequence $x$ of length $L_x$, let $R_{x,t}\in\mathbb{R}^{d\times d}$
denote the RoPE rotation at token position $t$, and define
\[
[\mathcal R_x(X)]_{t,:}
=
\left(R_{x,t}X_{t,:}^{\top}\right)^{\top},
\qquad
H_{L_x}
=
I_{L_x}
-
\frac{1}{L_x}
\mathbf{1}_{L_x}\mathbf{1}_{L_x}^{\top},
\]
The key reconstruction error is
\begin{equation}
e_{\ell,r}^{K}
=
\frac{
\displaystyle
\sum\nolimits_{x\in\mathcal D_{\rm cal}}
\left\|
H_{L_x}
\mathcal R_x\!\left(
X_{\ell,x}^{K}
-
\widehat X_{\ell,r,x}^{K}
\right)
\right\|_F^2
}{
\displaystyle
\sum\nolimits_{x\in\mathcal D_{\rm cal}}
\left\|
H_{L_x}
\mathcal R_x\!\left(
X_{\ell,x}^{K}
\right)
\right\|_F^2
}.\nonumber
\end{equation}
Here $H_{L_x}$ removes the mean over token positions after RoPE, so the
key error is normalized by the variation that remains visible to attention
rather than by a large common component.

SAKV compares these errors with separate thresholds $\varepsilon_K$ and
$\varepsilon_V$. Given $\varepsilon_\tau$, the retained rank of group
$\mathcal L$ is the smallest rank for which every layer in the group
satisfies the corresponding error limit,
\begin{equation}
r_{\mathcal L}^{\tau}
=
\min
\left\{
r:
e_{\ell,r}^{\tau}
\leq
\varepsilon_{\tau}
\quad
\text{for every } \ell\in\mathcal L
\right\},
\qquad
\tau\in\{K,V\}.
\label{eq:sakv-rank}
\end{equation}
The thresholds $\varepsilon_K$ and $\varepsilon_V$ are determined from the
target compression ratio as described below.

\paragraph{Layer grouping.}
For a group $\mathcal L$, using rank $r_{\mathcal L}^{\tau}$ requires
\begin{equation}
C_{\mathcal L}^{\tau}
=
Lr_{\mathcal L}^{\tau}
+
|\mathcal L|r_{\mathcal L}^{\tau}d\nonumber
\end{equation}
stored values. The first term stores the shared matrix $A_{\mathcal L}^{\tau}$ and the second stores the layer-specific $B_{\ell}^{\tau}$ matrices. Layers that share a compact spectral space satisfy
Eq.~\ref{eq:sakv-rank} at a smaller rank. Less compatible layers require more rank and increase the cache cost. SAKV selects a contiguous partition
by
\begin{equation}
\mathcal P^{*}
=
\arg\min_{\mathcal P}
\sum\nolimits_{\mathcal L\in\mathcal P}
\left(
C_{\mathcal L}^{K}
+
C_{\mathcal L}^{V}
\right),
\qquad
|\mathcal L|\leq g_{\max},
\label{eq:sakv-partition}
\end{equation}
where $\mathcal P$ partitions the Transformer layers into contiguous groups.
Eq.~\ref{eq:sakv-partition} is solved by dynamic programming.

The thresholds $\varepsilon_K$ and $\varepsilon_V$ are determined separately for each target compression ratio. For trial threshold values, Eq.~\ref{eq:sakv-rank} Eq.~\ref{eq:sakv-partition}  determine the required ranks and layer partition. Bisection adjusts $\varepsilon_K$ and $\varepsilon_V$ until the resulting configuration reaches the target compression. The calibration stage then returns one shared layer partition and one key and one value rank for each group. These groups and ranks are fixed after calibration.

During inference, SAKV uses the layer groups and ranks selected during calibration. For each group $\mathcal L$ and cache type $\tau$, the rank $r_{\mathcal L}^{\tau}$ selected by Eq.~\ref{eq:sakv-rank} is used in the reconstruction of Eq.~\ref{eq:sakv-reconstruction}. SAKV changes only the
layer grouping and rank allocation, while retaining the pretrained model and
the xKV cache reconstruction procedure.

\paragraph{Results.}
We evaluate SAKV on Llama-3.1-8B-Instruct \cite{llama3} and Qwen2.5-14B-Instruct-1M \cite{qwen25_1m} using RULER \cite{ruler} at context lengths of 32K, 64K, and 128K. We compare with xKV \cite{xkv} at the same compression ratios using the released xKV rank configurations. Detailed RULER results, paired statistical tests, LongBench \cite{longbench} results, and ablations are reported in
Appendix~\ref{app:sakv}.

The main advantage of SAKV appears at high compression. At approximately $4\times$ compression, SAKV and xKV perform similarly. As the compression ratio increases, xKV degrades much faster, while SAKV retains substantially more accuracy. On Llama-3.1-8B at 64K, xKV scores 88.50 at \(8\times\) compression but drops to 63.17 at \(12\times\) and 35.10 at \(16\times\). SAKV retains 85.48 and 79.66 at the latter two compression ratios. In particular, SAKV at \(12\times\) remains only 3.02 points below xKV at \(8\times\), whereas xKV loses 25.33 points when moving from \(8\times\) to \(12\times\).

The same pattern appears on Qwen2.5-14B. At 64K, SAKV improves over xKV by 4.64 points at $12\times$ compression and 18.92 points at $16\times$. Across all 20 matched RULER settings, SAKV is significantly better in 14 and tied in 6. The gains are concentrated at the higher compression ratios, where adaptive layer grouping and rank allocation matter most.

\subsection{CAPS: Channel-Aligned Pruning Selection}
\label{sec:caps}

Pruning aims to reduce computation while preserving model performance. Unstructured pruning can flexibly remove individual weights, but the resulting irregular sparsity is difficult to execute efficiently on standard hardware. Structured pruning instead removes predefined weight units, such as channels, neurons, attention heads, or weight blocks, which makes the computation reduction easier to realize. However, removing all weights associated with a larger unit together usually leads to greater performance loss, since some of these weights may still be important to the model
output.

Existing structured-pruning methods mainly differ in how they estimate the importance of the structured units. Wanda-sp \cite{wanda}, FLAP \cite{flap}, RCPU \cite{rcpu}, and LLM-Pruner \cite{llmpruner}, for example, use weight, activation, or dependency information to decide which units should be removed. Better importance estimation can improve which unit is selected, but the weights within each selected unit are still removed together. Consider input-channel pruning. An input channel may be important to some output neurons but contribute little to others. Removing the entire channel also removes the useful connections.

CAPS uses a finer structured pruning unit. It first divides the output neurons into groups according to their GSA structure. For each group, CAPS keeps only the input channels that the group relies on and removes the remaining connections for that group. The same input channel can therefore be kept for some groups and pruned for others. Because all neurons in a group use the same retained input channels, the corresponding computation can still be carried out as a dense operation on that smaller input set. CAPS therefore removes unnecessary computation more selectively while preserving a structure that can be efficiently executed.

\paragraph{Method.}
Consider the weight matrix of a linear layer 
\(
W\in\mathbb{R}^{d_{\mathrm{out}}\times d_{\mathrm{in}}}\) and its singular value decomposition $W = U\Sigma V^\top$, where each row of $W$ corresponds to an output neuron and each column corresponds to an input channel. CAPS first determines how many singular directions are needed to describe the main spectral structure of this particular layer. We compute the effective rank
\[
R_{\varepsilon}(W)
=
\min
\left\{
R:
\frac{\sum_{j=1}^{R}\sigma_j^2(W)}
{\sum_j\sigma_j^2(W)}
\geq 1-\varepsilon
\right\},
\]
as defined in Eq.~\ref{eq:effective-rank}. We then form
\begin{equation}
P
=
U_{:,1:R_{\varepsilon}(W)}\Sigma_{1:R_{\varepsilon}(W),1:R_{\varepsilon}(W)}.\nonumber
\end{equation}
The first $R_{\varepsilon}(W)$ singular directions are used only to organize the output neurons into groups. Row $r$ of $P$ gives the coordinates of output neuron $r$ along these directions. 

For each of these $R_{\varepsilon}(W)$ directions, CAPS selects the output neurons with the largest absolute coordinates as anchor candidates. Candidates with highly similar weight vectors are removed as redundant, leaving a set of distinct anchors that serve as the group centers. Each remaining output neuron is then assigned to the anchor with which its weight vector has the largest absolute
cosine similarity. This produces groups $\mathcal G_1,\ldots,\mathcal G_K$, each centered on one of the distinct anchors.

CAPS next determines which input channels each group should keep. We run the pretrained model on a small calibration set and let $a_c$ denote the $\ell_2$ norm of the activation of an input channel $c$. For group $\mathcal G_i$, let $W[\mathcal G_i,c]$ denote the vector of weights connecting input channel $c$ to all output neurons in $\mathcal G_i$. CAPS calculates the score of input channel $c$ in the group as
\begin{equation}
s_i(c)
=
\left\|
W[\mathcal G_i,c]
\right\|_2 a_c .\nonumber
\end{equation}
The first term measures the total weight strength from input channel $c$ to the output neurons in $\mathcal G_i$. The second term measures how strongly that channel is used on the calibration data, following the weight-times-activation criterion used by Wanda \cite{wanda}.

For a score-retention threshold $\rho$, CAPS ranks channels in $\mathcal G_i$ by $s_i(c)$ and keeps the smallest set $\mathcal C_i(\rho)$ such that the sum of the selected channel scores is at least a fraction $\rho$ of the total channel score:
\begin{equation}
\sum\nolimits_{c\in\mathcal C_i(\rho)}s_i(c)
\geq
\rho
\sum\nolimits_{c=1}^{d_{\mathrm{in}}}s_i(c).\nonumber
\end{equation}
The channel selection is performed separately for every $\mathcal G_i$, so different output-neuron groups in the same layer can keep different input channels.

If output neuron $r$ belongs to group $\mathcal G_i$, CAPS keeps its connections from the input channels in $\mathcal C_i(\rho)$ and removes its connections from the other input channels. All output neurons in the same group use the same selected input channels, while different groups can use different channel sets. For group $\mathcal G_i$, the remaining computation involves only weight parameters
\(
W[\mathcal G_i,\mathcal C_i(\rho)].
\)

For the linear layer, the original matrix multiplication requires computation
proportional to
\(
d_{\mathrm{out}}d_{\mathrm{in}}.
\)
After CAPS pruning, the computation is proportional to
\begin{equation}
\sum\nolimits_{i=1}^{K}
|\mathcal G_i|\,|\mathcal C_i(\rho)|.\nonumber
\end{equation}
Each group is evaluated with a smaller dense matrix multiplication, reducing the group-wise MAC count. The realizable hardware reduction depends on how groups with different input-channel sets are tiled. Appendix~\ref{app:caps-compute} analyzes this effect. At the same time, an input channel does not need to be removed from the entire layer, giving CAPS finer pruning flexibility than whole-channel pruning.

For a convolutional layer with weight tensor
\(W\in\mathbb{R}^{C_{\mathrm{out}}\times C_{\mathrm{in}}\times k\times k}\), CAPS first reshapes $W$ into \(C_{\mathrm{out}}\times (C_{\mathrm{in}}k^2)\), so that each row contains all kernel weights of one output channel. The same SVD-based anchor grouping is then applied to the rows of reshaped weights to obtain groups $\mathcal G_1,\ldots,\mathcal G_K$.

The input-channel selection is performed based on original weights $W$. For group $\mathcal G_i$ and
input channel $c$, let \(W[\mathcal G_i,c,:,:]\) denote all $k^2$ kernel weights connecting input channel $c$ to the output channels in $\mathcal G_i$. CAPS scores input channel $c$ for this group as
\begin{equation}
s_i(c)
=
\left\|
W[\mathcal G_i,c,:,:]
\right\|_F a_c,\nonumber
\end{equation}
where $a_c$ is the $\ell_2$ norm of the activation of input channel $c$ on the calibration data. The Frobenius norm measures the total weight strength between input channel $c$ and all output channels in the group, including all $k^2$ spatial kernel coefficients.

As in the linear case, CAPS ranks the input channels by $s_i(c)$ and selects the smallest set $\mathcal C_i(\rho)$. For removed channel $c\notin\mathcal C_i(\rho)$, all kernel weights $W[o,c,:,:]$ are removed for every output channel $o\in\mathcal G_i$. The computation for group $\mathcal G_i$ is reduced from being proportional to $|\mathcal G_i|C_{\mathrm{in}}k^2$ to
$|\mathcal G_i||\mathcal C_i(\rho)|k^2$. The resulting computation remains a dense convolution
over a smaller set of input channels and can be executed efficiently using standard dense kernels.

\paragraph{Results.}
We evaluate CAPS on Llama-2-7B~\cite{llama2}, Llama-3-8B~\cite{llama3}, Mistral-7B~\cite{jiang_mistral_2023}, and Qwen2.5-7B~\cite{qwen25} at target MAC reductions of $10\%$, $20\%$, and $30\%$. We prune the seven linear projections in each transformer block and compare with magnitude pruning, Wanda-sp \cite{wanda}, FLAP \cite{flap}, and RCPU \cite{rcpu}. All methods use the same calibration data and receive comparable recovery training after pruning. We evaluate WikiText-2 perplexity, average accuracy over seven zero-shot tasks (ZS-7), and 5-shot MMLU.

CAPS obtains the lowest WikiText-2 perplexity at all 12 model--budget settings, the highest ZS-7 accuracy at 11 of 12 settings, and the highest MMLU accuracy at 10 of 12 settings. The differences are largest on the three GQA models. At $20$--$30\%$ MAC reduction, CAPS improves over the strongest
competing method by $0.50$--$1.01$ perplexity, $2.94$--$7.29$ ZS-7 points, and $7.59$--$14.60$ MMLU points. The gains on Llama-2-7B are smaller, but CAPS still gives lower perplexity at all three pruning levels. Full results are reported in Appendix~\ref{app:caps}.

\section{Related Work}
\label{sec:related}

This section reviews prior work most closely related to the CHASE applications, covering PEFT, structured pruning and compensation, model merging, attention reconfiguration, and KV-cache compression.

\paragraph{Parameter-efficient finetuning.}
LoRA represents task adaptation as an additive low-rank update to a frozen pretrained weight \cite{lora}. AdaLoRA allocates the low-rank budget across weight matrices during training rather than using the same rank everywhere \cite{adalora}. DoRA separates the magnitude and direction of a weight before applying low-rank adaptation \cite{dora}, while PiSSA initializes the low-rank factors from the principal singular components of the pretrained weight \cite{pissa}. OFT updates pretrained weights through orthogonal transformations \cite{oft}, and BOFT reduces the parameter cost of orthogonal finetuning through butterfly factorization \cite{boft}. CHASE CORA \cite{cora} differs from these methods by coupling the rotations of the left and right singular bases and allowing the singular values to change separately. Its parameterization follows the coherent basis motion analyzed by GSA-II rather than treating the update as an unrestricted low-rank matrix or a one-sided orthogonal transformation.

\paragraph{Structured pruning and post-pruning compensation.}
Unstructured pruning removes individual weights and provides fine-grained control over which parameters are discarded. SparseGPT uses approximate second-order reconstruction to select individual weights for removal \cite{sparsegpt}, while Wanda combines weight magnitude with activation statistics to score individual weights \cite{wanda}. Although these methods can preserve performance well at high sparsity, the resulting irregular sparsity is difficult to translate into real acceleration on standard hardware.

Structured pruning instead removes predefined groups of weights, such as channels, neurons, attention heads, or weight blocks. LLM-Pruner uses gradient information to identify coupled structures for removal \cite{llmpruner}, ZipLM selects structured components according to their effect on loss and runtime cost \cite{ziplm}, and Sheared LLaMA learns a smaller architecture by removing layers, heads, and hidden dimensions \cite{shearedllama}. FLAP performs training-free channel pruning using activation statistics \cite{flap}. Removing larger structures makes the computation reduction easier to realize in hardware, but all weights within the selected unit are removed together, including connections that may still be important to part of output neurons. CHASE CAPS uses the leading singular directions selected by the effective-rank criterion to group output neurons. It then selects the retained input channels separately for each group using weight and activation information. An input channel may be kept for one group and removed for another. This gives CAPS a finer pruning granularity while keeping the computation within each group dense, so that the reduction in computation can be realized more efficiently. The actual MAC reduction depends on how these groups are mapped to row tiles in the implementation.

Post-pruning compensation addresses the distortion introduced after the pruning structure has been fixed. FLAP compensates the mean output error caused by channel removal with a bias term \cite{flap}, while RCPU applies an orthogonal transformation and scaling to the retained weights \cite{rcpu}. These
corrections recover only limited aspects of the pruned representation. Structured pruning can change both the singular values of the retained matrix and the orientations of its input and output subspaces, causing a mismatch with the original layer that cannot be corrected by an output bias or a single transformation. CHASE COEC \cite{coec} addresses this problem by jointly correcting the left and right singular bases of the retained matrix and adjusting its singular values. The compensation preserves the reduced model dimensions and requires no retraining, while recovering more of the geometry lost during pruning.

\paragraph{Model merging.}
Model merging combines several models fine-tuned from the same pretrained base model without jointly retraining them. Model Soups averages independently fine-tuned weights \cite{modelsoups}, while Fisher merging weights model parameters using estimates of their Fisher information \cite{fishermerge}. Task Arithmetic represents each fine-tuned model by its difference from the pretrained model and combines these task vectors in weight space \cite{taskarithmetic}. TIES removes small updates and resolves sign conflicts before merging \cite{ties}, and DARE randomly removes and rescales entries of the task vectors before applying a merging rule \cite{dare}. These methods mainly manipulate parameters or parameter updates in the original weight coordinates. However, fine-tuning can change not only individual weights but also the singular values and singular subspaces of a layer. Directly combining the resulting updates can mix changes that correspond to different geometric directions and increase interference between models. OrthoMerge reduces such interference through an SVD-based orthogonalization of model updates \cite{orthomerge}. CHASE CORAM \cite{coram} takes a different approach. It expresses each fine-tuned weight relative to the corresponding pretrained weight through rotational, spectral, and right-subspace changes, and merges these components separately. This allows changes in singular values and subspace orientations to be combined according to their own geometry rather than treating the entire fine-tuning update as a single weight-space displacement.

\paragraph{MHA-to-GQA conversion.}
Multi-query attention (MQA) reduces decoding memory by sharing one set of key and value projections across all query heads \cite{mqa}. Grouped-query attention (GQA) provides an intermediate design in which several query heads share each key/value head \cite{gqa}. Converting a pretrained multi-head
attention (MHA) model to GQA requires both deciding which query heads should share a KV head and constructing the shared key and value projections. The original GQA conversion groups existing heads and initializes each shared projection by directly averaging the corresponding MHA projections before
continued training.

Later methods improve the conversion by changing how heads are grouped or aligned before merging. AsymGQA forms nonuniform groups using activation similarity~\cite{asymgqa}. The work of~\cite{mha2gqa_align} aligns attention heads through Procrustes transformations while preserving the original computation, and then removes redundant KV heads through $\ell_0$-regularized pruning. These methods improve the compatibility of the heads being combined, but the shared key and value projections are still formed by combining multiple pretrained heads. Direct averaging can discard
head-specific directions, while alignment before averaging only reduces the coordinate mismatch and does not recover information lost when several value projections are replaced by one.

CAGA addresses the grouping problem using the GSA structure of the pretrained heads. For each pair of heads, it measures physical-channel support overlap, retained singular-value similarity, and retained right-subspace similarity. These quantities are combined to form the head-pair distance used to construct equal-size GQA groups.

CAGA separately constructs the shared key and value projections within each group. The key projections are aligned to a shared key projection using RoPE-compatible transformations, with the head-specific transformations absorbed into the corresponding query projections. For the values, CAGA stacks the grouped value projections and extracts a shared rank-$d_h$ subspace. The head-specific factors are absorbed into the output projections. The resulting layer has the standard GQA form with one shared key/value head per group and no additional inference-time transformations.

\paragraph{KV-cache compression.}
The key-value (KV) cache is a major source of memory consumption during autoregressive decoding, particularly for long contexts and deep models. Existing methods reduce this cost through low-rank compression, token selection, or sharing across layers. Palu compresses cached keys and values
through low-rank representations \cite{palu}, while ShadowKV exploits low-dimensional structure in the cache to reduce the amount of stored KV information \cite{shadowkv}. MiniCache reduces cache redundancy across nearby layers \cite{minicache}, and cross-layer attention shares KV representations
across layers by changing the attention architecture \cite{cla}. More recently, xKV constructs shared low-rank representations for multiple layers through aligned singular-vector extraction \cite{xkv}. These methods show that substantial redundancy exists not only within individual KV caches but
also across layers. CHASE SAKV extends the GSA effective-rank criterion to grouped KV representations. It evaluates the reconstruction error of candidate contiguous layer groups as the retained rank varies, selects separate key and value ranks under their respective tolerances, and chooses the contiguous partition accordingly. The grouping and rank allocation therefore follow the measured spectral structure of the cached representations rather than a fixed sharing pattern and fixed ranks.

\section{Conclusion}
\label{sec:conclusion}

GSA-I and GSA-II identify spectral concentration, physical-channel alignment, support structure, and singular-basis changes in trained networks. In this paper, we propose CHASE to use these structures in model modification, reconfiguration, and compression.

We present six CHASE applications. CORA uses coherent singular-basis changes for PEFT. COEC uses two-sided geometric compensation and spectral adjustment after structured pruning. CORAM separates rotational, spectral, and subspace changes when merging finetuned models. CAGA  uses GSA to identify multi-head attention heads that can share a KV representation and constructs the shared key and value heads through geometric alignment and low-rank subspace extraction. SAKV uses GSA to determine which adjacent layers can share a low-rank KV-cache representation and the retained rank for each layer group. CAPS uses GSA spectral structure to group output neurons and selects retained input channels separately for each group. 

These applications demonstrate the structures identified by GSA can be used directly to design methods for a range of model operations. Results from CORA, COEC, and CORAM have already shown the effectiveness of GSA for PEFT, pruning compensation, and model merging. In the new applications, CAGA shows that geometric shared-head construction substantially improves MHA-to-GQA conversion, while SAKV and CAPS improve over their respective baselines for KV-cache compression and structured pruning.


\appendix

\section{Detailed Experiments for CAGA}
\label{app:CAGA}

\paragraph{Model and conversion settings.}
We use Llama-2-7B \cite{llama2}, which has $H=32$ query heads in each attention layer. We evaluate conversion to $G=16$ and $G=8$ key/value heads, corresponding to $2\times$ and $4\times$ reductions in the number of KV heads. 

\paragraph{Calibration data.}
We run the pretrained model on 128 C4 training sequences of 2,048 tokens and collect the hidden states entering each attention layer. These samples are used to compute the head statistics, determine the head groups, and estimate the key-alignment transformations. The evaluation sets are not used during conversion.

\paragraph{Grouping methods.}
We compare three head-grouping methods. Neighboring-head grouping places adjacent query heads in the same group, following the standard GQA conversion \cite{gqa}. The group heads of~\cite{mha2gqa_align} use their cosine/value-based criterion. 

\paragraph{Shared-head construction.}
We compare direct averaging, the Procrustes-based conversion of~\cite{mha2gqa_align}, and the CAGA construction described in Sec.~\ref{sec:caga}. CAGA aligns the key projections using the
RoPE-compatible transformations and constructs the shared value projection from the rank-$d_h$ approximation of the grouped value matrices.

\subsection{Effect of Shared-Head Construction}

We first compare different ways of constructing the shared key and value heads before any continued training. Table~\ref{tab:caga-construction} reports the results at $G=16$.

Directly averaging independently trained key and value projections loses a large amount of information. Aligning the heads before averaging improves the result, while the CAGA construction gives a much larger improvement. Without any continued training, CAGA reduces WikiText-2 perplexity from 926 to 25.1 and increases the seven-task zero-shot average from 32.6 to 44.3.

\begin{table}[t]
\centering
\small
\caption{Llama-2-7B at $G=16$ before continued training. Results use the
CAGA head grouping.}
\label{tab:caga-construction}
\begin{tabular}{l r r}
\toprule
Shared-head construction & WikiText-2 PPL & ZS-7 \\
\midrule
Direct averaging & 926 & 32.6 \\
Procrustes alignment & 738 & 35.3 \\
CAGA & \textbf{25.1} & \textbf{44.3} \\
\bottomrule
\end{tabular}
\end{table}

\subsection{Grouping Comparison and Continued Training}

We compare CAGA with neighboring-head grouping and the cosine/value-based
grouping of~\cite{mha2gqa_align}. All three methods use the same
CAGA shared-head construction, so the comparison only changes how the query
heads are grouped.

For the $G=8$ model, we perform continued training on the same SlimPajama data \cite{slimpajama} in the same order. Training uses full-parameter updates with AdamW, a learning rate of $2\times10^{-5}$, weight decay of 0.1, 5\% warmup, and cosine learning-rate decay. Each update contains 128 sequences of 2,048 tokens. Knowledge distillation from the original Llama-2-7B model is used during training.

We evaluate perplexity on WikiText-2 \cite{wikitext} and C4 \cite{c4}. The seven-task zero-shot average includes BoolQ, RTE, HellaSwag, WinoGrande, ARC-Easy, ARC-Challenge, and OpenBookQA. Table~\ref{tab:caga-grouping-training} reports the results after 100M and 500M training tokens.

\begin{table}[t]
\centering
\small
\caption{Llama-2-7B at $G=8$ after continued training. All models use the
same CAGA shared-head construction.}
\label{tab:caga-grouping-training}
\begin{tabular}{l r r r r r r}
\toprule
& \multicolumn{3}{c}{100M tokens}
& \multicolumn{3}{c}{500M tokens} \\
\cmidrule(lr){2-4}\cmidrule(lr){5-7}
Grouping & WT2 & C4 & ZS-7 & WT2 & C4 & ZS-7 \\
\midrule
Neighboring heads
& 7.78 & 9.57 & 50.39
& 6.75 & 8.66 & 53.52 \\
Jin et al.
& \textbf{7.55} & \textbf{9.35} & 51.03
& \textbf{6.64} & \textbf{8.55} & 53.49 \\
CAGA
& 7.67 & 9.47 & \textbf{51.58}
& 6.68 & 8.60 & \textbf{54.36} \\
\bottomrule
\end{tabular}
\end{table}

The three grouping methods perform similarly at both training budgets. After 500M tokens, WikiText-2 perplexity ranges from 6.64 to 6.75 and the zero-shot average ranges from 53.49 to 54.36. The choice of grouping has a much smaller effect than the construction of the shared key and value heads.

\section{Additional Evaluation of SAKV}
\label{app:sakv}

\subsection{Experimental Setup}

\paragraph{Models and evaluation.}
We evaluate SAKV on Llama-3.1-8B-Instruct \cite{llama3} and Qwen2.5-14B-Instruct-1M \cite{qwen25_1m}. Long-context performance is measured on RULER \cite{ruler} at context lengths of 32K, 64K, and 128K.
For each setting, xKV \cite{xkv} and SAKV are evaluated on the same prompts. We consider approximately $4\times$, $8\times$, $12\times$, and $16\times$ KV-cache compression when supported by the corresponding model and context length. SAKV and xKV are compared at the same compression ratio.

We additionally evaluate both methods on LongBench \cite{longbench} at $4\times$ and $8\times$ compression to examine whether the behavior observed on RULER transfers to a broader set of long-context tasks.

\paragraph{Calibration.}
SAKV uses four long sequences from PG-19 \cite{rae2020compressive} for calibration. For each candidate contiguous layer group, we compute the reconstruction error over candidate ranks and
select the layer partition and key/value ranks using Eqs.~\ref{eq:sakv-rank}--\ref{eq:sakv-partition}. We use $g_{\max}=8$. The key and value thresholds $\varepsilon_K$ and $\varepsilon_V$ are determined separately for each target compression ratio. The resulting groups and ranks are fixed for
evaluation. No model parameters are trained or modified during calibration.

\subsection{Comparison with xKV}

Table~\ref{tab:sakv-ruler} reports the RULER results. At $4\times$ compression, both methods remain close to the uncompressed model and differ little from each other. The advantage of SAKV becomes more obvious as the compression ratio increases.

On Llama-3.1-8B at 64K, xKV scores 88.50 at $8\times$ compression but drops to 63.17 at $12\times$ and 35.10 at $16\times$. SAKV retains 85.48 and 79.66 at the $12\times$ and $16\times$ compression ratios. In particular, SAKV at $12\times$ remains only 3.02 points below xKV at $8\times$, whereas xKV
itself loses 25.33 points when moving from $8\times$ to $12\times$. Similar behavior appears at 32K and 128K.

Qwen2.5-14B shows a similar trend. At 64K, the difference between SAKV and xKV increases from 2.77 points at $8\times$ to 4.64 points at $12\times$ and 18.92 points at $16\times$. SAKV provides its
largest advantage at high compression ratios where a fixed layer grouping and fixed rank allocation begin to lose substantial long-context accuracy.

\begin{table*}[t]
\centering
\caption{RULER average under matched KV-cache compression. ``Full" means
full KV-cache.}
\label{tab:sakv-ruler}
\small
\setlength{\tabcolsep}{5pt}
\begin{tabular}{llrrrrr}
\toprule
Model & Context & Full & Compression & xKV & SAKV & Gain \\
\midrule
Llama-3.1-8B
& 32K  & 93.63 & $4\times$  & 93.00 & 92.58 & -0.42 \\
&      &       & $8\times$  & 92.24 & 92.48 & +0.24 \\
&      &       & $12\times$ & 78.57 & 89.83 & +11.26 \\
&      &       & $16\times$ & 46.57 & 79.89 & +33.32 \\
\cmidrule(lr){2-7}
& 64K  & 92.10 & $4\times$  & 91.16 & 91.22 & +0.06 \\
&      &       & $8\times$  & 88.50 & 89.71 & +1.21 \\
&      &       & $12\times$ & 63.17 & 85.48 & +22.31 \\
&      &       & $16\times$ & 35.10 & 79.66 & +44.56 \\
\cmidrule(lr){2-7}
& 128K & 84.82 & $4\times$  & 81.43 & 82.11 & +0.68 \\
&      &       & $8\times$  & 72.69 & 76.87 & +4.18 \\
&      &       & $12\times$ & 43.53 & 68.62 & +25.09 \\
&      &       & $16\times$ & 21.87 & 56.61 & +34.74 \\
\midrule
Qwen2.5-14B
& 32K  & 94.38 & $4\times$  & 94.29 & 94.31 & +0.02 \\
&      &       & $8\times$  & 91.49 & 92.85 & +1.36 \\
\cmidrule(lr){2-7}
& 64K  & 93.49 & $4\times$  & 91.73 & 92.41 & +0.68 \\
&      &       & $8\times$  & 87.49 & 90.26 & +2.77 \\
&      &       & $12\times$ & 78.35 & 82.99 & +4.64 \\
&      &       & $16\times$ & 54.97 & 73.89 & +18.92 \\
\cmidrule(lr){2-7}
& 128K & 90.42 & $4\times$  & 87.34 & 89.45 & +2.11 \\
&      &       & $8\times$  & 80.22 & 84.52 & +4.30 \\
\bottomrule
\end{tabular}
\end{table*}

Because xKV and SAKV are evaluated on the same 960 RULER prompts, we compute the per-prompt SAKV–xKV score difference and form 95\% confidence intervals using 10,000 paired bootstrap resamples. SAKV is significantly better in 14 of the 20 matched settings, statistically indistinguishable in 6, and worse in none. The indistinguishable cases occur primarily at $4\times$ compression, where both methods remain close to the uncompressed model.

\subsection{LongBench Performance}

Table~\ref{tab:sakv-longbench} reports the LongBench results. At the $4\times$ and $8\times$ compression ratios, both xKV and SAKV remain close to the uncompressed model. Unlike RULER at higher
compression, these settings do not produce a clear separation between the two methods.

\begin{table}[t]
\centering
\caption{Average LongBench score.}
\label{tab:sakv-longbench}
\small
\begin{tabular}{lrrrr}
\toprule
Model & Full & Compression & xKV & SAKV \\
\midrule
Llama-3.1-8B
& 41.46 & $4\times$ & 40.84 & 41.26 \\
&       & $8\times$ & 40.06 & 40.20 \\
\midrule
Qwen2.5-14B
& 46.66 & $4\times$ & 46.41 & 46.39 \\
&       & $8\times$ & 46.59 & 45.96 \\
\bottomrule
\end{tabular}
\end{table}

\subsection{Additional Analysis of SAKV}

\paragraph{Layer grouping versus rank allocation.}
We first separate the effect of layer grouping from that of adaptive rank selection. At 64K and $8\times$ compression, using the SAKV groups with the xKV ranks gives 89.81 on Llama-3.1-8B and 90.18 on Qwen2.5-14B, compared with 88.50 and 87.49 for xKV. At this compression ratio, much of the improvement already comes from replacing the fixed four-layer grouping. At higher compression, rank allocation becomes increasingly important because the available rank must be distributed more selectively across depth.

\paragraph{Separate key and value tolerances.}
Keys and values exhibit different spectral decay, making one reconstruction threshold unsuitable for both cache types. On Llama-3.1-8B at 64K and $8\times$ compression, using one shared tolerance gives a RULER score of 10.31, compared with 89.62 when $\varepsilon_K$ and $\varepsilon_V$ are selected separately. 

\paragraph{RoPE-aware key criterion.}
The standard relative Frobenius error is sufficient when the spectral energy of a key representation reflects the variation relevant to attention. This assumption does not hold for the early layers of Qwen2.5-14B, where a large token-independent component dominates the key energy. Under the Frobenius
criterion, these layers can receive extremely small ranks even though the remaining variation becomes important after RoPE. Table~\ref{tab:sakv-rope} compares the two criteria on Qwen2.5-14B at 64K. The difference is small at $8\times$, but becomes large at $12\times$ and $16\times$. 

\begin{table}[t]
\centering
\caption{Effect of the key reconstruction criterion on Qwen2.5-14B at 64K.}
\label{tab:sakv-rope}
\small
\begin{tabular}{lrr}
\toprule
Compression & Frobenius & RoPE-aware \\
\midrule
$8\times$  & 90.94 & 90.26 \\
$12\times$ & 59.58 & 82.99 \\
$16\times$ & 55.74 & 73.89 \\
\bottomrule
\end{tabular}
\end{table}

The RoPE-aware criterion is not uniformly better than the Frobenius criterion. For example, on Llama-3.1-8B at 32K and $16\times$ compression, the Frobenius criterion obtains 89.36 compared with 79.89 for the RoPE-aware criterion. We use the latter because a single criterion must remain reliable across both model families, particularly in the settings where offset-dominated key representations cause the Frobenius criterion to fail.

\paragraph{Maximum group size.}
The maximum allowed group size controls how broadly SAKV may share one spectral basis across neighboring layers. On Llama-3.1-8B at 64K and $8\times$ compression, increasing $g_{\max}$ from 2 to 4 and 8 improves the RULER score from 84.37 to 87.56 and 89.62, respectively. We use $g_{\max}=8$ throughout the main experiments.

\paragraph{Shared and separate key/value partitions.}
SAKV uses one layer partition for keys and values while allowing their ranks to differ. Allowing the two cache types to choose separate partitions gives little additional benefit. At 64K and $8\times$ compression, the shared and separate variants obtain 89.71 and 89.75 on Llama-3.1-8B, and 90.26
and 90.30 on Qwen2.5-14B. 

\paragraph{Calibration length.}
The grouping and rank allocation are not sensitive to matching the calibration length exactly to the evaluation length. For Llama-3.1-8B evaluated at 64K and $8\times$ compression, calibrating at 32K, 64K, and 128K gives RULER scores of 89.57, 89.62, and 89.00, respectively.

\section{Additional Experiments for CAPS}
\label{app:caps}

\paragraph{Models.}
We evaluate CAPS on Llama-2-7B~\cite{llama2}, Llama-3-8B~\cite{llama3}, Mistral-7B~\cite{jiang_mistral_2023}, and Qwen2.5-7B~\cite{qwen25}. For every transformer block, pruning is applied to the seven linear projections $W_Q$, $W_K$, $W_V$, $W_O$, $W_{\rm gate}$,
$W_{\rm up}$, and $W_{\rm down}$. Embeddings, normalization layers, and the language-model head are left unchanged.

\paragraph{CAPS configuration.}
For each target layer, CAPS computes the anchor groups from the dense pretrained weight matrix. We set $\varepsilon=0.4$, so the singular directions containing $60\%$ of the spectral energy are used for grouping. For each direction, the four output neurons with the largest absolute coordinates are selected as anchor candidates. Candidate anchors whose weight vectors have an absolute cosine similarity above $0.85$ are treated as redundant. Each remaining output neuron is assigned to the non-redundant anchor with the largest absolute cosine similarity.

For each target pruning level, we search for the value of $\rho$ (defined in Sec.\ref{sec:caps}) that produces the desired MAC reduction. Since decreasing $\rho$ keeps fewer input channels, we use bisection to adjust $\rho$ until the MAC reduction computed from the selected channel sets $\{\mathcal C_i(\rho)\}$ is within $0.2$ percentage points of the target. The anchor groups are computed once from the dense model. After pruning to the target MAC reduction, the model receives one recovery-training phase while the removed connections remain fixed at zero.

\paragraph{Calibration.}
All methods use the same WikiText-2 training data~\cite{wikitext} for calibration. We sample 128 windows of 4096 tokens with seed 0, giving 524,288 calibration tokens. For CAPS, the activation term $a_c$ is computed as
\[
a_c=
\sqrt{\sum\nolimits_t x_{t,c}^2},
\]
where $x_{t,c}$ is the activation of input channel $c$ at token $t$.

\paragraph{Recovery training.}
We use the same SlimPajama~\cite{slimpajama} token stream for CAPS and other methods. The recovery budget is proportional to the target MAC reduction. The $10\%$, $20\%$, and $30\%$ settings use 34.1M, 66.6M, and 99.9M tokens, respectively. Training uses AdamW with learning rate $2\times10^{-5}$, $\beta=(0.9,0.95)$, weight decay $0.1$, $5\%$ warmup, and cosine decay to $10\%$ of the peak learning rate. We use sequences of length 2048 and an effective batch size of 128 sequences per optimization step. The Llama-2-7B baselines at the $30\%$ setting use 100.9M recovery tokens, approximately $1\%$ more than CAPS. All other comparisons use the same recovery budget at a given target.

\paragraph{Baselines.}
We compare CAPS with structured magnitude pruning, Wanda-sp \cite{wanda}, FLAP \cite{flap}, and RCPU \cite{rcpu}. These methods use whole-unit structured pruning, where a selected channel, neuron, or attention unit is removed together with all weights associated with that unit, and the
corresponding matrix dimension is reduced. In contrast, CAPS removes an input channel only for selected output-neuron groups and retains it for other groups.

For Llama-3-8B, Mistral-7B, and Qwen2.5-7B, which use GQA, the baseline attention layers are pruned at KV-group granularity. Their achieved MAC reductions can therefore differ slightly from the nominal target. CAPS instead adjusts $\rho$ so that its group-wise pruning reaches the target MAC reduction
directly.

\paragraph{Evaluation.}
We evaluate language modeling using WikiText-2 test perplexity with sequence length 4096. ZS-7 is the arithmetic mean of BoolQ, RTE, HellaSwag, WinoGrande, ARC-Easy, ARC-Challenge, and OpenBookQA. We use normalized accuracy for HellaSwag, ARC-Easy, ARC-Challenge, and OpenBookQA, and accuracy for the other three tasks. For Llama-3-8B and Qwen2.5-7B, the evaluation harness uses the first 2000 examples for tasks containing more than 2000 examples. All comparisons are made within the same model and therefore use the same evaluation set. MMLU~\cite{mmlu} is evaluated 5-shot over all 57 subjects using the development examples as demonstrations. We report macro accuracy over the 57 subjects. Each model--method--budget configuration is run once. Calibration uses seed 0 and recovery training uses seed 1234 with the same data order for all methods.

\subsection{Results}
\label{app:caps-results}

Table~\ref{tab:caps-full} reports the  comparison after recovery training. CAPS obtains the lowest WikiText-2 perplexity at all 12 model--budget settings. It obtains the highest ZS-7 accuracy at 11 of the 12 settings and the highest MMLU accuracy at 10 of the 12 settings.

The differences are particularly large on the three GQA models. At $20$--$30\%$ MAC reduction on Llama-3-8B, Mistral-7B, and Qwen2.5-7B, CAPS reduces perplexity by $0.50$--$1.01$ relative to the strongest baseline for that metric. The corresponding improvements reach $2.94$--$7.29$ points on ZS-7 and $7.59$--$14.60$ points on MMLU. On Llama-2-7B, differences in ZS-7 and MMLU are smaller, while CAPS retains lower perplexity at all three pruning levels.

\begin{table*}[t]
\centering
\caption{CAPS results after recovery training. Each method entry reports PPL / ZS-7 / MMLU. The MAC column gives the measured MAC reduction for CAPS and for the corresponding whole-unit baselines, respectively.}
\label{tab:caps-full}
\scriptsize
\setlength{\tabcolsep}{3.2pt}
\resizebox{\textwidth}{!}{
\begin{tabular}{llcccccc}
\toprule
Model & Target &
MAC $\downarrow$ CAPS / base &
CAPS &
Magnitude &
Wanda-sp &
FLAP &
RCPU \\
\midrule
Llama-2-7B
& 10\% & 9.83 / 9.79
& 5.2119 / 63.23 / 44.00
& 7.7365 / 58.51 / 24.78
& 5.3761 / 63.00 / 41.71
& 5.3932 / 62.06 / 42.49
& 5.3797 / 62.26 / 40.40 \\
& 20\% & 19.94 / 19.59
& 5.4775 / 61.14 / 39.37
& 10.0504 / 53.94 / 26.00
& 5.7983 / 59.43 / 36.43
& 5.8373 / 61.27 / 37.61
& 5.8405 / 60.82 / 39.79 \\
& 30\% & 29.91 / 30.41
& 5.9217 / 59.91 / 33.41
& 12.4877 / 48.06 / 23.75
& 6.4232 / 55.92 / 32.61
& 6.4348 / 56.05 / 31.72
& 6.4240 / 57.39 / 34.19 \\
\midrule
Llama-3-8B
& 10\% & 10.03 / 10.48
& 6.2232 / 66.54 / 64.55
& 9.5231 / 59.49 / 50.40
& 6.8051 / 65.55 / 57.50
& 7.1379 / 65.64 / 58.45
& 6.8218 / 65.94 / 58.67 \\
& 20\% & 19.94 / 20.96
& 6.7981 / 64.87 / 61.38
& 15.1003 / 54.36 / 25.20
& 7.8110 / 60.46 / 46.78
& 8.1345 / 58.84 / 46.55
& 7.8754 / 60.38 / 46.72 \\
& 30\% & 30.15 / 29.04
& 7.6100 / 62.13 / 54.18
& 22.3106 / 49.26 / 26.19
& 8.4428 / 57.12 / 41.01
& 8.7243 / 58.02 / 41.69
& 8.4145 / 59.19 / 41.36 \\
\midrule
Mistral-7B
& 10\% & 10.17 / 10.48
& 5.3543 / 67.23 / 58.40
& 6.8511 / 62.14 / 47.33
& 5.6296 / 64.68 / 55.59
& 5.6283 / 63.50 / 53.97
& 5.6130 / 64.33 / 53.33 \\
& 20\% & 19.81 / 20.96
& 5.5443 / 65.45 / 57.51
& 8.0033 / 57.64 / 36.24
& 6.0930 / 58.23 / 44.15
& 6.2721 / 57.68 / 42.34
& 6.1082 / 58.06 / 45.58 \\
& 30\% & 30.08 / 29.04
& 5.8704 / 64.23 / 53.26
& 8.8854 / 56.22 / 28.63
& 6.3904 / 56.94 / 36.71
& 6.5447 / 56.68 / 40.56
& 6.3722 / 56.15 / 36.94 \\
\midrule
Qwen2.5-7B
& 10\% & 10.01 / 8.74
& 6.5235 / 69.72 / 74.84
& 7.9283 / 66.53 / 70.75
& 6.6607 / 69.37 / 72.52
& 6.6691 / 69.45 / 72.76
& 6.6574 / 68.71 / 72.81 \\
& 20\% & 20.07 / 20.63
& 6.8264 / 69.18 / 73.05
& 11.4809 / 59.63 / 59.27
& 7.6786 / 64.30 / 64.85
& 7.7317 / 64.84 / 65.46
& 7.5681 / 62.27 / 62.39 \\
& 30\% & 30.12 / 29.37
& 7.3083 / 67.28 / 69.44
& 13.1102 / 55.09 / 53.06
& 8.1237 / 61.30 / 59.83
& 8.2299 / 60.84 / 60.06
& 7.9831 / 60.09 / 56.83 \\
\bottomrule
\end{tabular}}
\end{table*}

\subsection{Comparison with the Best Whole-Unit Baseline}

Table~\ref{tab:caps-delta} reports the difference between CAPS and the strongest whole-unit baseline for each metric. Since the strongest baseline can differ across metrics, PPL, ZS-7, and MMLU are compared independently. For $\Delta$PPL, a negative value denotes lower perplexity for CAPS. For
$\Delta$ZS-7 and $\Delta$MMLU, a positive value denotes higher accuracy.

\begin{table}[t]
\centering
\caption{Difference between CAPS and the strongest whole-unit baseline for
each metric. Lower $\Delta$PPL is better. Higher $\Delta$ZS-7 and $\Delta$MMLU are better.}
\label{tab:caps-delta}
\small
\setlength{\tabcolsep}{4pt}
\begin{tabular}{lrrrr}
\toprule
Model & Target & $\Delta$PPL & $\Delta$ZS-7 & $\Delta$MMLU \\
\midrule
Llama-2-7B   & 10\% & -0.16 & +0.23 & +1.51 \\
              & 20\% & -0.32 & -0.13 & -0.42 \\
              & 30\% & -0.50 & +2.52 & -0.78 \\
\midrule
Llama-3-8B   & 10\% & -0.58 & +0.60 & +5.88 \\
              & 20\% & -1.01 & +4.41 & +14.60 \\
              & 30\% & -0.80 & +2.94 & +12.49 \\
\midrule
Mistral-7B   & 10\% & -0.26 & +2.55 & +2.81 \\
              & 20\% & -0.55 & +7.22 & +11.93 \\
              & 30\% & -0.50 & +7.29 & +12.70 \\
\midrule
Qwen2.5-7B   & 10\% & -0.13 & +0.27 & +2.03 \\
              & 20\% & -0.74 & +4.34 & +7.59 \\
              & 30\% & -0.67 & +5.98 & +9.38 \\
\bottomrule
\end{tabular}
\end{table}

\subsection{Computation Accounting}
\label{app:caps-compute}

For CAPS, the reported MAC reduction is computed directly from the selected input channels of the output-neuron groups. For a pruned linear layer, the remaining multiply-add count is proportional to
\[
\sum\nolimits_i |\mathcal G_i|\,|\mathcal C_i|.
\]
The reported MAC reduction is the reduction in this count relative to the dense versions of the seven target linear layers. For the whole-unit baselines, the MAC count is computed from the dimensions of the reduced matrices.

The CAPS computation is exact when each output-neuron group is evaluated using its own selected input channels. On current accelerator hardware, however, dense matrix kernels operate on row tiles that may contain several CAPS groups. If rows with different selected input channels are evaluated in the same tile, the tile must use the union of their input-channel sets. The realizable MAC reduction therefore depends on the row granularity supported by the implementation.

To quantify this effect, we reorder rows according to the overlap of their selected input channels and combine them into tiles of $T$ rows. For each tile, computation is counted over the union of the input channels used by its rows. Reordering does not change the model output. Table~\ref{tab:caps-tiles} reports the resulting MAC reduction. At the nominal $30\%$ setting, a tile size of two rows retains $24.2$--$24.5\%$ MAC reduction across the four models. The reduction becomes smaller as the tile size increases because more rows with different selected input channels must share the same input set. Qwen2.5-7B retains the largest reduction at larger tile sizes.

\begin{table*}[t]
\centering
\caption{MAC reduction of CAPS under different row-tile sizes. $T=1$
corresponds to the group-wise MAC count used in the pruning experiments.
For $T>1$, rows are reordered according to overlap of their selected input
channels before tiling.}
\label{tab:caps-tiles}
\small
\setlength{\tabcolsep}{5pt}
\begin{tabular}{llrrrrrr}
\toprule
Model & Target & $T=1$ & $T=2$ & $T=4$ & $T=8$ & $T=32$ & $T=128$ \\
\midrule
Llama-2-7B
& 10\% & 9.83 & 6.59 & 3.41 & 1.80 & 0.72 & 0.39 \\
& 20\% & 19.94 & 15.13 & 9.01 & 4.88 & 1.78 & 0.94 \\
& 30\% & 29.91 & 24.33 & 15.98 & 8.94 & 2.97 & 1.43 \\
\midrule
Llama-3-8B
& 10\% & 10.03 & 6.67 & 3.67 & 1.92 & 0.64 & 0.32 \\
& 20\% & 19.94 & 15.03 & 9.33 & 5.06 & 1.50 & 0.62 \\
& 30\% & 30.15 & 24.49 & 16.68 & 9.55 & 2.73 & 1.02 \\
\midrule
Mistral-7B
& 10\% & 10.17 & 6.72 & 3.82 & 2.06 & 0.81 & 0.53 \\
& 20\% & 19.81 & 14.77 & 9.22 & 4.95 & 1.52 & 0.81 \\
& 30\% & 30.08 & 24.21 & 16.46 & 9.24 & 2.55 & 1.16 \\
\midrule
Qwen2.5-7B
& 10\% & 10.01 & 6.85 & 4.70 & 3.13 & 1.75 & 1.34 \\
& 20\% & 20.07 & 15.14 & 10.79 & 6.95 & 2.90 & 1.87 \\
& 30\% & 30.12 & 24.40 & 18.31 & 12.13 & 4.57 & 2.56 \\
\bottomrule
\end{tabular}
\end{table*}





\end{document}